\documentclass{article} % For LaTeX2e
\usepackage{collas2022_conference,times}

\usepackage{amsmath,amsfonts,bm}

\newcommand{\newterm}[1]{{\em #1}}

\def\eqref#1{equation~\ref{#1}}
\def\1{\bm{1}}

\def\vo{{\bm{o}}}

\def\vs{{\bm{s}}}

\def\vv{{\bm{v}}}
\def\vw{{\bm{w}}}

\def\vomega{{\boldsymbol{\omega}}}
\def\vnu{{\boldsymbol{\nu}}}

\def\evs{{s}}

\def\evv{{v}}

\DeclareMathAlphabet{\mathsfit}{\encodingdefault}{\sfdefault}{m}{sl}
\SetMathAlphabet{\mathsfit}{bold}{\encodingdefault}{\sfdefault}{bx}{n}

\newcommand{\Ls}{\mathcal{L}}

\newcommand{\softmax}{\mathrm{softmax}}

\newcommand{\sigmoid}{\sigma}

\newcommand{\mgd}{\textbf{MGD}}
\newcommand{\obs}{\textbf{obs}}
\newcommand{\expert}{\textbf{exp}}

\newcommand{\FAlin}{\texttt{lin}}
\newcommand{\FAagg}{\texttt{agg}}

\usepackage{hyperref}
\hypersetup{
    colorlinks=true,
    linkcolor=red,
    filecolor=magenta,      
    urlcolor=blue,
    citecolor=purple,
    pdftitle={Overleaf Example},
    pdfpagemode=FullScreen,
    }

\usepackage{caption}
\usepackage{subcaption}
\usepackage{graphicx}
\usepackage{algorithm}
\usepackage{algpseudocode}
\usepackage{wrapfig}

\title{What Should I Know? Using Meta-gradient Descent for Predictive Feature Discovery in a Single Stream of Experience}

\author{
Alexandra Kearney \thanks{hi@alexkearney.com}, Anna Koop \thanks{akoop@ualberta.ca}\\
Department of Computing Science\\
University of Alberta\\
Edmonton, AB, Canada
\And
Johannes G\"unther \\
Alberta Machine Intelligence Institute \&\\ 
Department of Computing Science\\
University of Alberta\\
Edmonton, AB, Canada\\
\AND
Patrick M. Pilarski\\
Alberta Machine Intelligence Institute \&\\
Department of Medicine \&\\
Department of Computing Science\\
University of Alberta\\
Edmonton, AB, Canada 
}

\collasfinalcopy 

\begin{document}

\maketitle

\begin{abstract}
In computational reinforcement learning, a growing body of work seeks to construct an agent's perception of the world through predictions of future sensations; predictions about environment observations are used as additional input features to enable better goal-directed decision-making. An open challenge in this line of work is determining from the infinitely many predictions that the agent could possibly make which predictions might best support decision-making. This challenge is especially apparent in continual learning problems where a single stream of experience is available to a singular agent. As a primary contribution, we introduce a meta-gradient descent process by which an agent learns 1) what predictions to make, 2) the estimates for its chosen predictions, and 3) how to use those estimates to generate policies that maximize future reward---all during a single ongoing process of continual learning. In this manuscript we consider predictions expressed as General Value Functions: temporally extended estimates of the accumulation of a future signal. We demonstrate that through interaction with the environment an agent can independently select predictions that resolve partial-observability, resulting in performance similar to expertly specified GVFs. By learning, rather than manually specifying these predictions, we enable the agent to identify useful predictions in a self-supervised manner, taking a step towards truly autonomous systems.
\end{abstract}

\section{Making Sense of the World Through Predictions}

It has long been suggested that predictions of future experience can provide useful and intuitive features to support decision-making---particularly in partially-observable or non-Markovian environments \citep{singh2003learning, littman2001predictive, jaeger2000observable}. It is certainly true for biological agents: humans and many animals build predictive sensorimotor models of their world. These predictions of experience form the basis for biological perception \citep{rao1999predictive, wolpert1995internal,gilbert2009stumbling}. A principled and well understood way of making temporally extended predictions in computational reinforcement learning is by estimating many value functions. Value functions predict the long-term expected accumulation of a signal in a given state \citep{sutton1988learning}, and can predict not only reward, but any signal available to an agent via its senses \citep{white2015developing}. Prior works have used these General Value Function (GVF) estimates as features to adapt the control interfaces of bionic limbs \citep{edwards2016application}, design reflexive control systems for robots \citep{modayil2014prediction} and living cats \citep{dalrymple2020pavlovian}, and to inform industrial welding systems of estimated weld quality \citep{GUNTHER20161}. In this paper, we explore how an agent can independently choose GVFs in order to augment its observations in order to construct it's own \textit{agent-state}: an approximation of state from the agent's subjective perspective.

An open challenge when using GVF estimates as input features is determining what aspects of an agent's experience to predict. Of all the possible predictions an agent could make, which subset of GVFs are most useful to inform and support decision making? This choice is often made by the system designer \citep{modayil2014prediction,dalrymple2020pavlovian,edwards2016application,GUNTHER20161}. However, recent work has explored how an agent might independently specify its own GVFs. In \cite{schlegel2018baseline} an agent randomly selects the parameters that define what aspect of the environment is being estimated. After a period of learning, the agent replaces a subset of its GVFs based on their learning progress. 

Determining which GVFs to replace is a core challenge for such {\em generate-and-test} approaches: A GVF may be accurate and have low prediction error, but just because a prediction is well estimated does not mean that it is useful as a predictive feature for control \citep{good-prediction}. Examining a learned prediction estimate without considering its use---as is done in generate-and-test for GVF specification---inherently limits the ability of an agent to choose {\em useful} predictive features.

An alternative to random selection of GVFs is to parameterise the specification of a GVF and perform meta-gradient descent.  By taking the gradient of a control learner's error with respect to a GVF's meta-parameters, what each prediction is about can be incrementally updated based on feedback from the control learner. Although not used for learning predictive inputs, recent work has shown early success in using meta-gradient descent as a means of learning meta-parameters that specify GVFs \citep{DBLP:journals/corr/abs-1909-04607} for use as auxiliary tasks \citep{jaderberg2016reinforcement}. 

When used as auxiliary tasks, GVF estimates themselves are not directly used in decision-making, but rather as regularisers for the control agent's artificial neural network. In this auxiliary task setting, the parameters that determine what is being predicted and the parameters that are used to select actions are explicitly kept and learned independent of one another. We propose meta update where the core RL update of a control learner directly influences {\em what} an agent is predicting.

\section{Meta-Learning General Value Functions}
       
In this manuscript, we integrate the discovery and use of GVFs for reinforcement learning control problems. We present a fully self-supervised approach, using meta-gradient descent to autonomously discover GVFs that are useful as predictive features for control. We do so by parameterising the functions that determine what aspect of the environment a GVF prediction is about, and constructing a loss that shapes the predictions based on the control agent's learning process. The resulting learning process can be successfully implemented incrementally and online. By this process, we enable agents to independently specify GVFs to be used directly as features by a control learner to solve two partially-observable problems. In doing so, we are providing a new solution to a long standing problem in using GVFs as predictive input features. 

\begin{figure}[h]
  \begin{subfigure}[t]{0.32\textwidth}
    \includegraphics[width=\linewidth,height=2in, keepaspectratio]{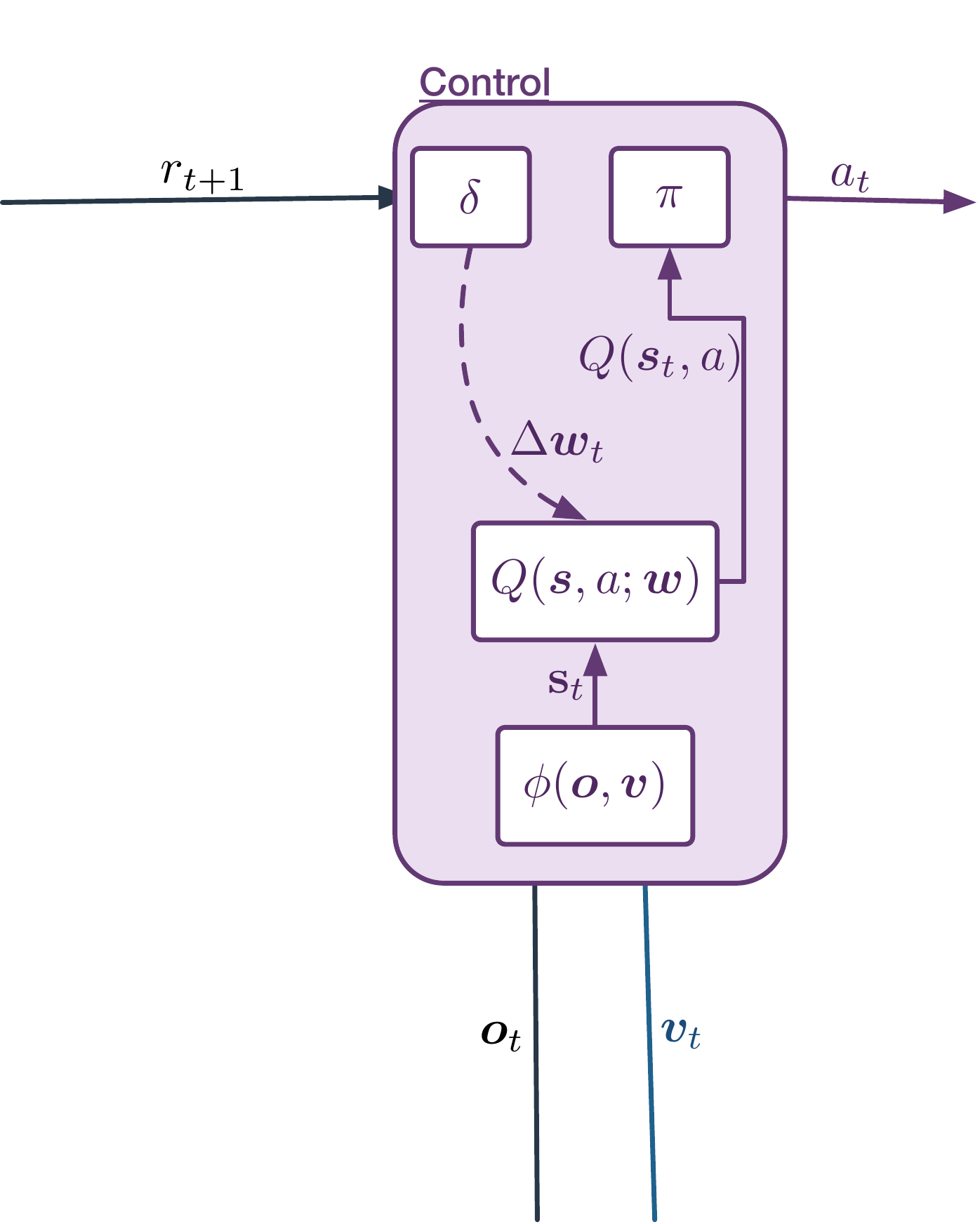}
    \centering
    \caption{The control agent updates its action-value function $Q(\vs, a; \vw)$ according to the TD-error $\delta$ and chooses actions according to policy $\pi$ based on value estimates..}
    \label{fig:control}
  \end{subfigure}\hfill % comment
  \begin{subfigure}[t]{0.32\textwidth}
        \includegraphics[width=\linewidth,height=2in, keepaspectratio]{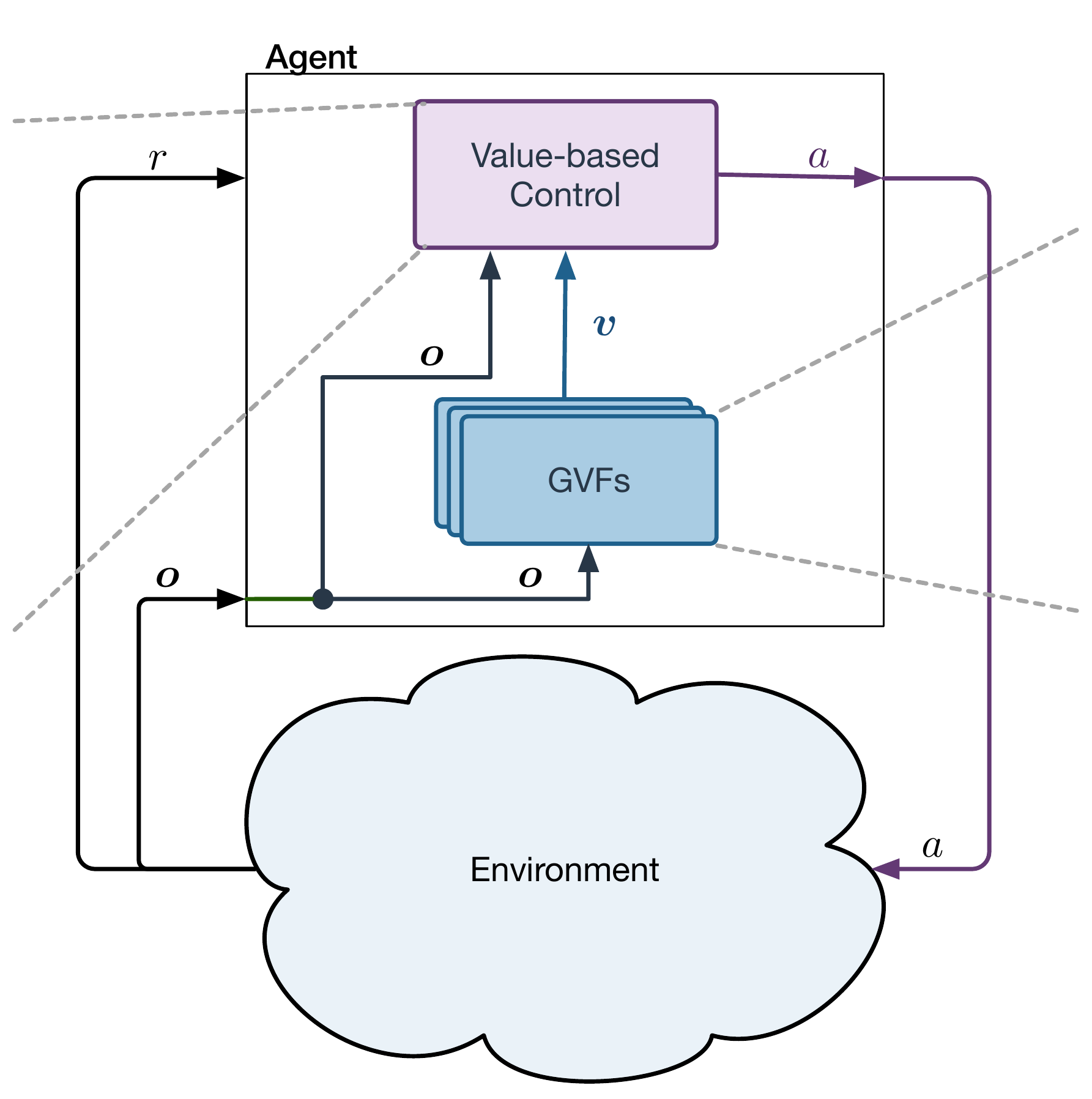}
         \centering
         \caption{The typical agent-environment diagram of Reinforcement Learning, where the control agent learns values as a function of both observations and GVF outputs.}
  \label{fig:architecture}
  \end{subfigure}\hfill % comment need for some obscure LaTeX reason
  \begin{subfigure}[t]{.32\textwidth}
    \includegraphics[width=\linewidth,height=2in, keepaspectratio]{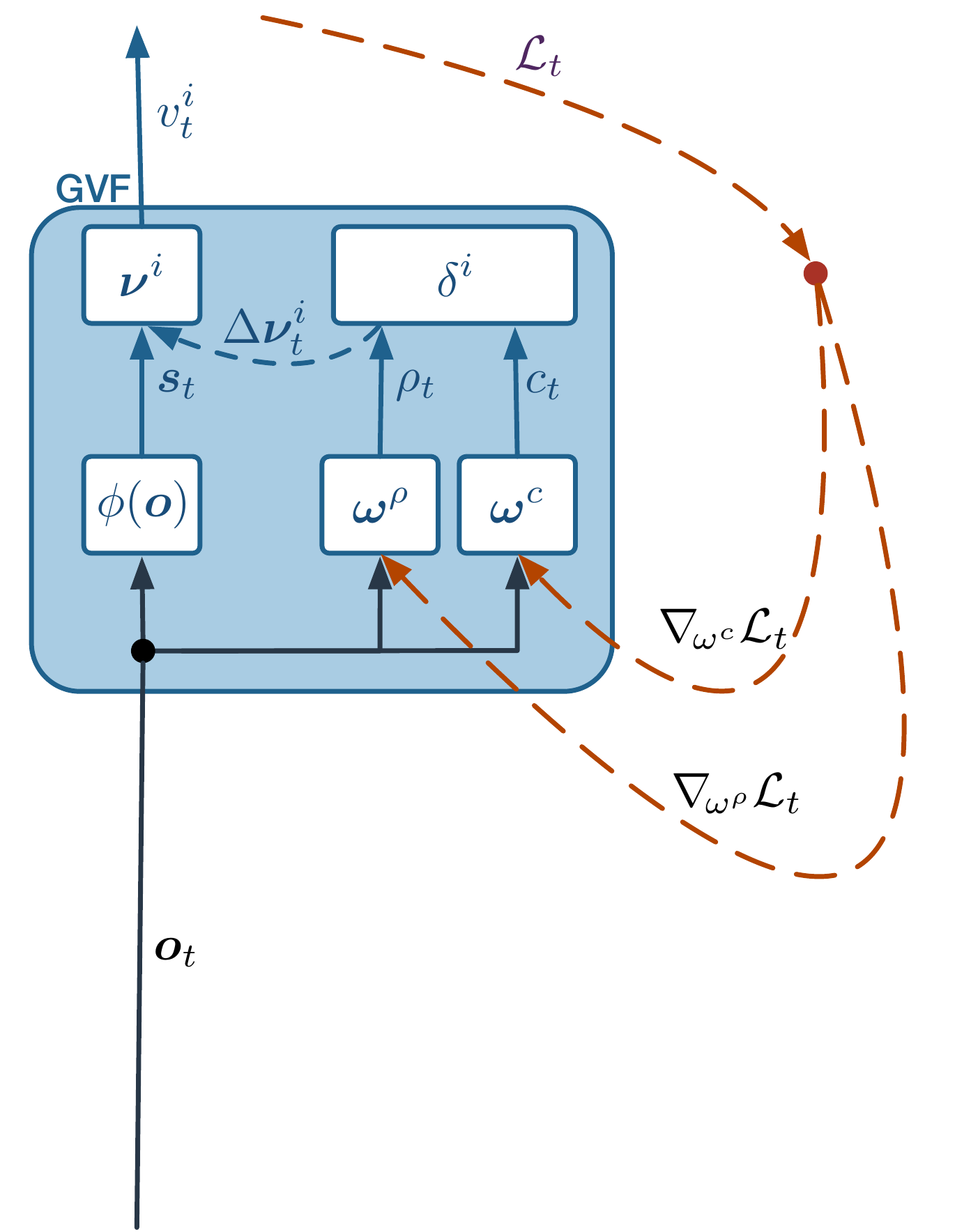}
    \centering
    \caption{A GVF outputs predictions according to its current weights $V(\vs; \vnu)$, while its updates are defined by the cumulant $c$, discount $\gamma$, and policy-correction $\rho$.}
    \label{fig:gvf}
    \end{subfigure}
    \caption{}
\end{figure}
\clearpage
Our meta-learning process (Figure \ref{fig:architecture}) operates on an agent structured in three parts : 1) a value-based control unit that learns weights $\vw$ for an action-value function (Figure~\ref{fig:control});
2) a collection of GVFs that each learn weights $\vnu$ to output prediction vector $\vv$ (Figure~\ref{fig:gvf}); and
%approximate value-functions at time $t$ specified by each GVF's current policy-recognizer $\rho$, cumulant $c$, and discount factor $\gamma$; and 
3) a set of meta-weights $\vomega$ that parameterize each GVF's learning rule (the right half of Figure~\ref{fig:gvf}). 

The control unit is a typical Q-learning agent, although it learns a value function over the \newterm{agent-state} $\vs$, which is constructed from the observations $\vo$ and a vector of GVF predictions $\vv$, rather than observation alone: 
\begin{equation}
\label{eq:agentstate}
    \vs_t = \phi(\vo_t, \vv_t).
\end{equation}
The action-value function $Q(\vs_t, a; \vw_t)$ may be learned by any relevant RL algorithm. We define the loss function $\Ls_t = \delta_t^2$, where:
\begin{equation}
\label{eq:delta}
     \delta_t = r_{t+1} + \gamma\max_a Q(\vs_{t+1}, a; \vw_t) - Q(\vs_t, a_t; \vw_t).
\end{equation}

GVFs may use any reinforcement learning method for learning, but their structure is defined by three functions of the current state\footnote{note that this is the GVF-specific state, independent of the \emph{agent-state} defined above}:  the cumulant or target $c$, the discount or termination signal $\gamma$, and the policy $\pi$ (see chapter 4 in \cite{white2015developing}).
The cumulant function $z(\vs_t)$ determines the current target $c_t$: in classic RL, the cumulant simply singles out the current reward $r_t$. The discount factor $\gamma$ determines how far into the future the signal-of-interest should be attended to. While in simple RL approaches it is generally a fixed value, in GVFs it can be any function of the current state $\gamma_t = g(\vs_t)$. The policy allows each GVF to condition its prediction on specific behaviours, and is used to compute the importance-sampling correction against the behaviour policy $\mu$: $\rho_t = \frac{\pi(\vs_t, a_t)}{\mu(\vs_t, a_t)}$. The output of a GVF is determined not only by the current weights $\vnu$ but also the functions that define its update procedure. We will use $\vomega$ to refer to the collective parameters that define the GVF structure, disambiguating with superscripts when necessary:
\begin{equation}
    v^i_t = V(\vs; \vnu^i, \vomega^i).
\end{equation}

During execution of our meta-learning process, each time-step contains an action and learning phase. First, the agent receives an observation from the environment $\vo_t$, which is used to compute the GVF state
\begin{equation}
    \vs^\nu_t = \phi^\nu(\vo_t).
\end{equation}
For each GVF $i$, the prediction value $v^i_t$ is calculated as a function of the GVF state $\vs^\nu_t$ and prediction weights $\vnu^i$:
\begin{equation}
    v^i_t = V(\vs^\nu_t; \vnu_t^i).
\end{equation}
The vector of GVF predictions, along with the current observation, is transformed into the agent-state using a fixed function (see Equation~\ref{eq:agentstate} where $\phi$ may include eg: state aggregation, tile coding, artificial neural net). The policy unit $\pi$ uses $Q(\vs_t, a; \vw_t)$ to determine the next action\footnote{in the following results we use $\epsilon$-greedy action selection}. Once the action is executed and $(\vo_{t+1}, r_{t+1})$ received, the learning phase begins.

The key to our meta-learning method is that the Q-learning error, as noted earlier (see Equation~\ref{eq:delta} and illustrated with the read lines in Figure~\ref{fig:gvf}), is a function of not only the value function weights $\vw$, but also the agent-state vector $\vs$.
As the agent-state is constructed from the GVF predictions, which in turn are are adjusted according to the meta-weights $\vomega$, we can use the control agent's loss function $\Ls_t=\delta^2_t$ to update the meta-weights. For the $i$th GVF, meta-weights $j\in\{c, \rho\}$ are adjusted:
\begin{equation}
    \Delta\vomega^{i,j}_t = \alpha^{i,j}\nabla_{\!\omega^{i,j}}\Ls_t
\end{equation}
Once the meta-weights have been updated, each GVF computes the current $\rho_t$, $c_t$, and $\gamma_t$, updates its predictions weights $\vnu$. Finally, updated agent-state $\vv_{t+1}$ is computed and used by the control agent to update its Q-learning weights $\vw_{t+1}$. Pseudocode is provided in Appendix \ref{appendix:alg}.

\section{Can An Agent Learn What To Predict?}

Using the meta process we introduced, can an agent find useful GVFs for use as predictive input features? We first evaluate meta specification of GVFs on a partially observable control problem, Monsoon World (Figure \ref{fig:monsoon}). We choose to introduce this problem as it is a minimal, clear example of a situation where temporal abstraction is neccessary to solve the problem. This enables us to assess in a clear way whether meta-gradient descent (MGD) is capable of specifying useful predictions, and precisely examine what predictions the agent chooses to learn. 

In Monsoon World, there are two seasons: monsoon and drought. The underlying season determines whether the agent receives reward for its chosen action, however the underlying season is not directly observable. Although the agent cannot directly observe seasons, it can observe the result of a given action: something impacted by the seasons. The agent tends to a field by choosing to either water, or not water their farm. Watering the field during a drought will result in a reward of 1; watering the field during monsoon season does not produce growth and results in a reward of 0, and vice versa during a monsoon. If the agent chooses the right action corresponding to the underlying season, a reward of 1 can be obtained on each time-step. Regardless of the action chosen by the agent, time progresses.

This monsoon world problem can be solved, and an optimal policy found, if the agent reliably estimates how long until watering produces a particular result. This can be done by learning \newterm{echo GVFs} (c.f. \citep{schlegel2021general}). Echo GVFs estimate the time to an event using a state-conditioned discount and cumulant. In plain terms, the two GVF's that when learned provide estimates that solve the problem can be described as ``How long until watering produces growth''? or ``How long until not watering produces growth''? Indirectly, these capture the time until either the monsoon or drought. These two predictions can be described as \newterm{off-policy} estimates: predictions that are conditioned on a particular behaviour. Given the agent has two actions where $a_0$ is not watering and $a_1$ is watering, we can describe the policy ``if the agent waters'' as a deterministic policy \(\pi = [0,1]\). The signal of interest is, \( c_t = 
      1 \text{ if } r_t = 1 \text{ \& } 0  \text{ otherwise}
\). Similarly, a state-dependent discounting function terminates the accumulation, \( \gamma_t =
      0  \text{ if } c_t = 1 \text{ \& } 0.9  \text{ otherwise.}
\) Off-policy GVFs can be estimated online, incrementally, while the agent is engaging in behaviours that do not strictly match the target policies of the prediction \citep{maei2011gradient}. \label{monsoon:intro} Having constructed the aforementioned GVFs, an agent can express what is hidden from its observation stream: how long until the next season. While no information was given about the season, by relating what is sensed by the agent with the actions that were taken, the agent is able to learn about the seasons indirectly.

\subsection{Learning to Specify GVFs in Monsoon World}

\begin{figure}
\begin{subfigure}[b]{0.4\textwidth}
    \includegraphics[width=\linewidth,height=1.5in,keepaspectratio]{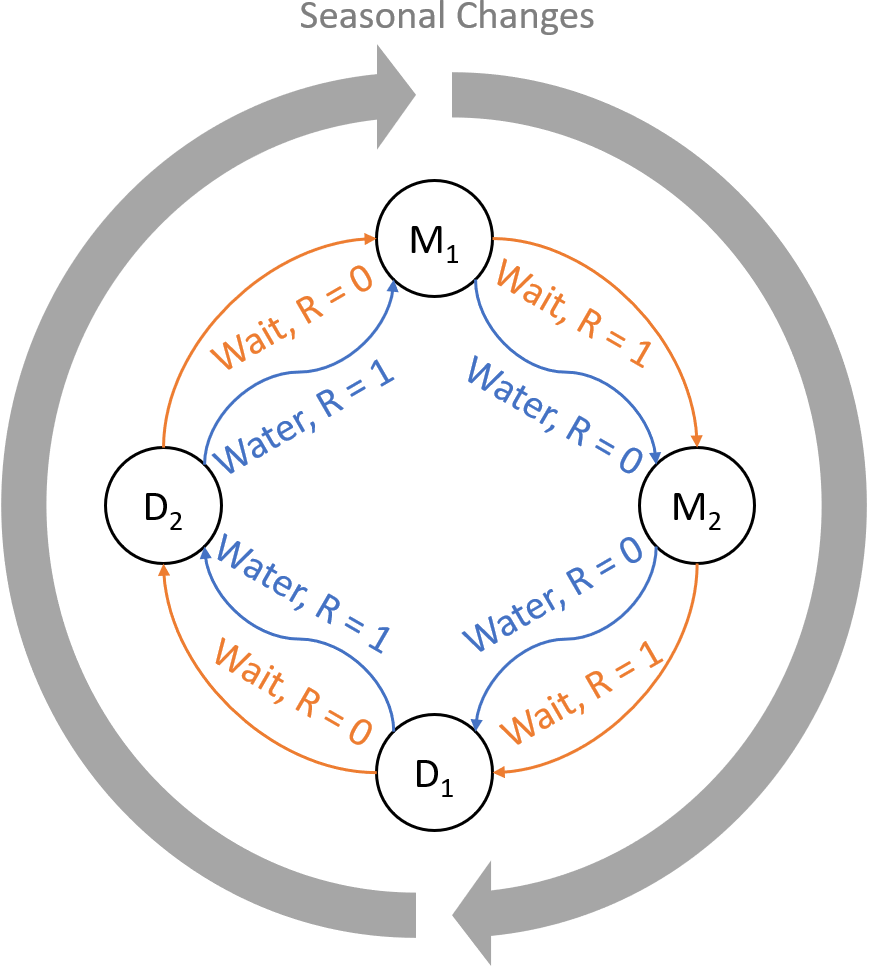}
    \centering
    \caption{The monsoon problem. On each time-step the agent observes a binary value that determines whether the crops have grown. Growth of crops is determined by both the action an agent takes (to water crops or not), and the unobserved underlying season.  There are four phases of the season that an agent can exist in: two monsoon and two drought (inner circles). The outer arrows indicate how the seasons change as the agent transitions through the cycle.}
    \label{fig:monsoon}
    \end{subfigure}
    \hfill
\begin{subfigure}[b]{0.55\textwidth}
\includegraphics[width=\linewidth, height=2in, keepaspectratio]{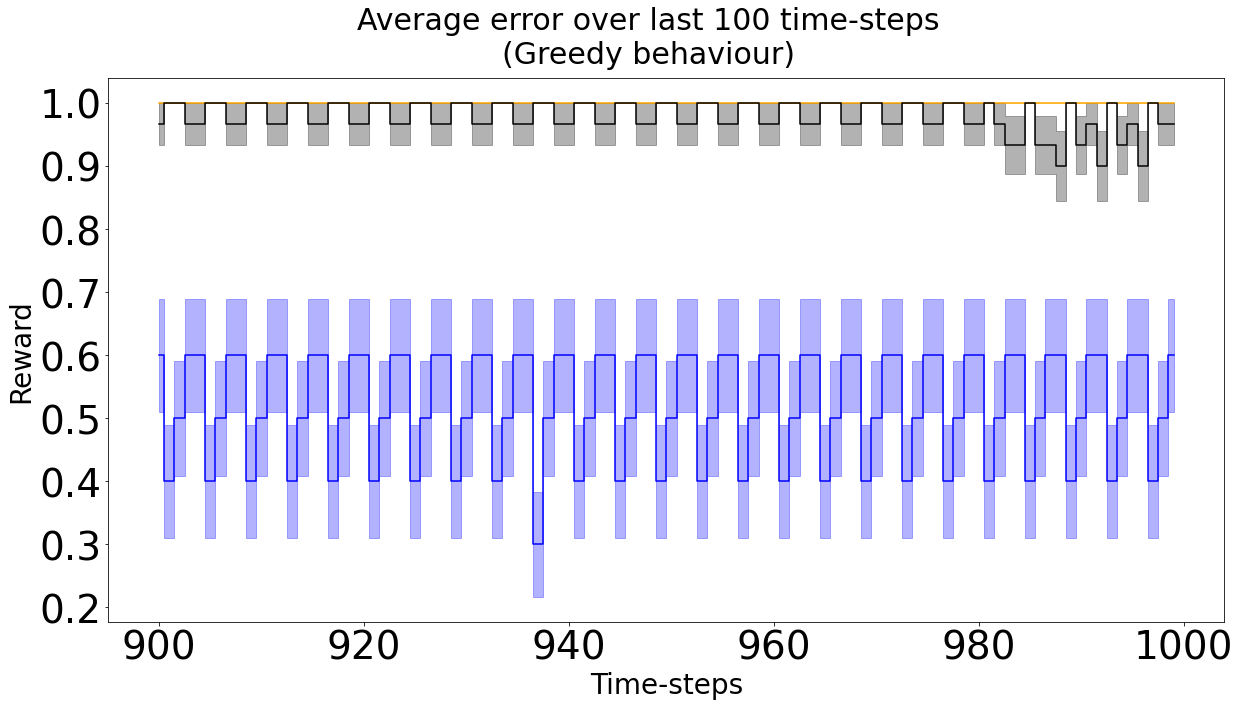}
\centering
\caption{Three different learners that use 1) the environmental observations as inputs (blue),  2) two additional inputs which express the seasons (in orange), 3) two additional predictions that are updated using meta-gradient descent (in black).  Each independent agent's mean reward is averaged over 30 independent trials. Error bars are standard error.}
\label{monsoon:results}
\end{subfigure}
\caption{}
\end{figure}

GVF estimates can resolve the partial observability of monsoon world. Through MGD, can an agent find similar predictions? We compare three different agent configurations (Figure \ref{monsoon:results}): 1) a baseline agent that only receives environmental observations as inputs (in blue), 2) an agent that in addition to the environmental observations, two inputs that capture underlying seasons (`oracle', in orange), and 3) an agent that has two GVFs whose cumulants and policies are learned through meta-gradient descent (in black).

Learning by MGD to specify GVFs introduces two additional sets of meta-weights to initialise: weights $\vomega^{\pi}$ that specify the policy a prediction is condition on, and weights $\vomega^{c}$ that determine the signal of interest from the environment that is being learnt about. Target policies are initialised  to an equiprobable weighting of actions, and are passed through a Softmax activation function $v_t^i = \mathtt{softmax}(\vo_t^\top \vw_t^{\pi})$ so that their sum is between [1, 0]. The cumulant meta-weights $\vomega^{c}$ are initialised to -5, and a sigmoid activation is applied such that $v^{c} = \mathtt{sigmoid}(\vo_t^\top \vomega^{c})$. We pass the cumulant meta-weights through a sigmoid to bound the cumulant between [0,1]. The meta-weights are updated each time-step incrementally. We apply an L2 regulariser to the meta-loss with $\lambda = 0.001$. The control learner is a linear Q-learner. Additional experiment details are in Appendix A.
        
In Figure \ref{monsoon:results}, we plot the average reward per time-step during the final 100 time-steps during which we evaluate agent performance given greedy behaviour. While the agents deterministically follow their policy during the evaluation phase, learning still occurs during the evaluation phase: updates are made to the GVFs, which affect the input observations to the control agent (in the case of the MGD agent), and the control learner continues to update its action-value function. This continued learning accounts for irregularity in the oscillations. 

The policy learnt using only environment observations is roughly equivalent to equiprobably choosing an action: the learned policy is no better than a coin-toss (Figure \ref{monsoon:results}, depicted in blue). This is as expected, given observations alone are insufficient to determine the optimal action on a given time-step. When the underlying season is provided as input (depicted in orange), the learned policy is approximately optimal. By augmenting environmental observations with predictive features that estimate the time to each season, an agent is able to solve the problem.  Using meta-gradient descent, the agent was able to select its own predictive features without any prior knowledge of the domain. {\em Using meta-gradient descent, the agent is able to solve the task with performance on-par with the hand-crafted solution without being given what to predict.}

\begin{figure}[t!]
\centering
\begin{subfigure}[b]{\linewidth}
\includegraphics[width=0.71\linewidth,height=3in, keepaspectratio]{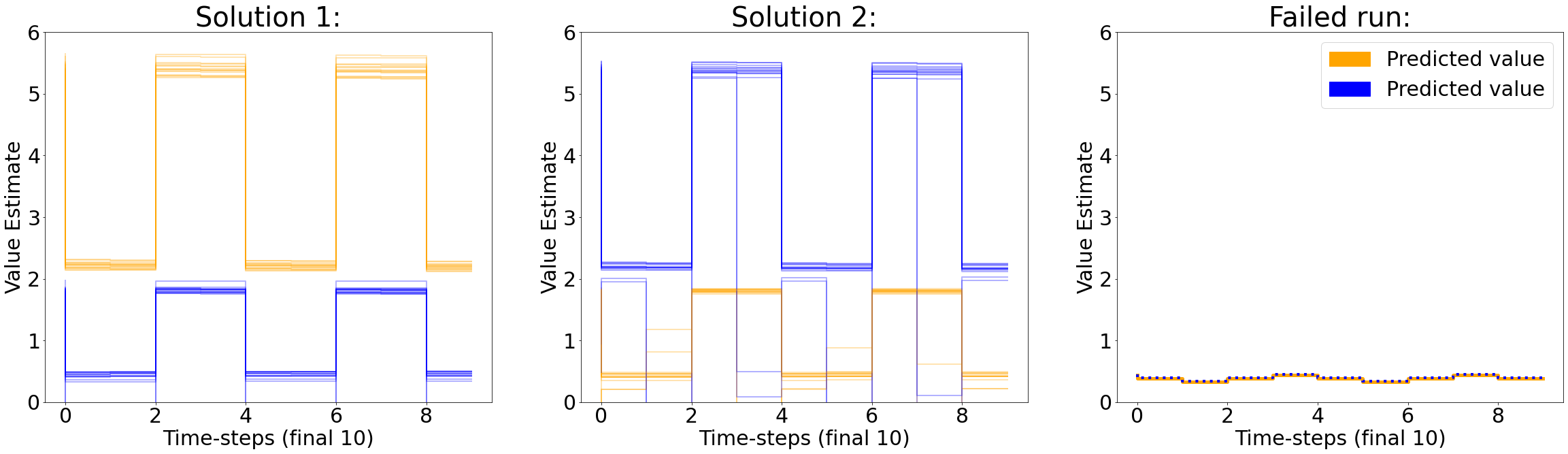}
\centering
\caption{Value estimates from each GVF during all independent trials.}
\label{fig:predictions}
\end{subfigure}
\vfill
\begin{subfigure}[b]{\linewidth}
\includegraphics[width=\linewidth,height=3in, keepaspectratio]{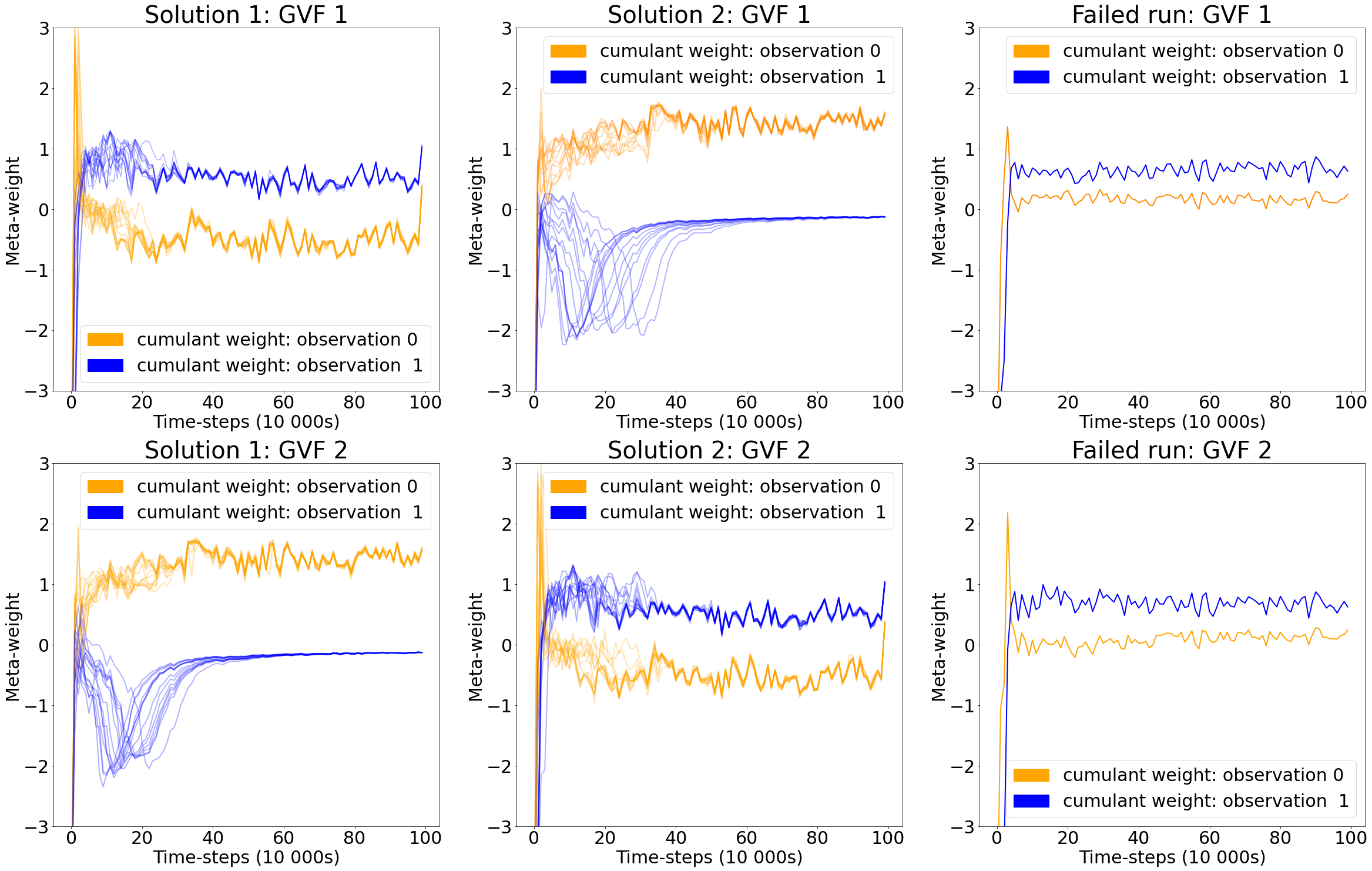}
\centering
\caption{The cumulant meta-weights $\vomega^c$ learned through MGD. }
\label{fig:cumulants}
\end{subfigure}
\vfill
\begin{subfigure}[b]{\linewidth}
\includegraphics[width=\linewidth,height=3in, keepaspectratio]{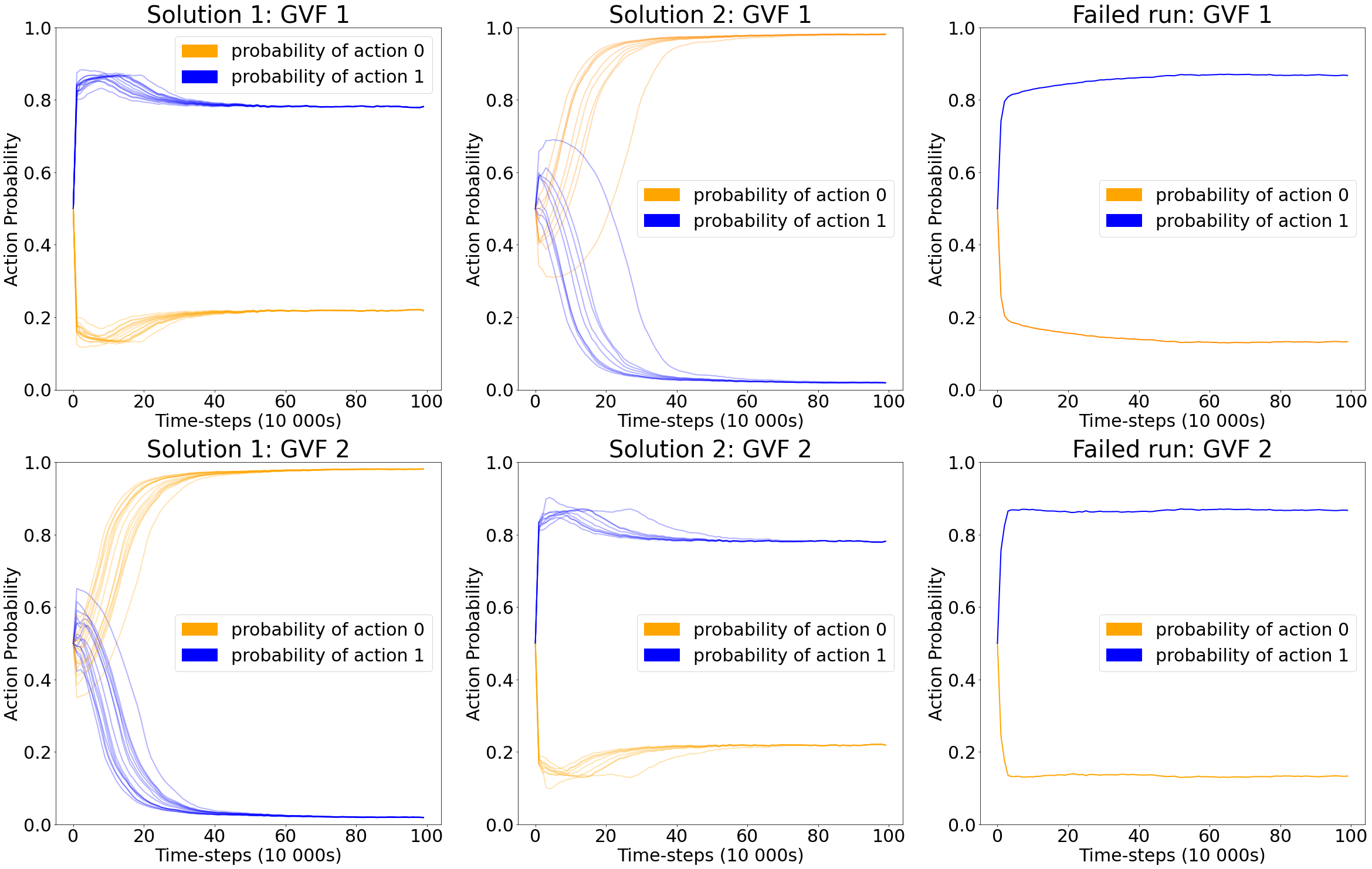}
\centering
\caption{The policy meta-weights $\vomega^\rho$ learned through MGD.}
\label{fig:policies}
\end{subfigure}
\caption{}
\label{fig:meta-params}
\end{figure}

If an agent can find GVFs to solve monsoon world, what are the useful predictions the agent found? Do they relate to our expert's GVFs? In Figure \ref{fig:predictions}, the value-estimates on the final 10 time-steps of each run are presented, as well as the meta-weights for the cumulants (Figure \ref{fig:cumulants}) and policies (Figure \ref{fig:policies}). We found that two distinct types of policy and cumulant were learned for each of the two GVFs. In one of the 30 independent trials, the agent failed to solve the problem, and we depict the learned meta-weights of this failed trial independently. The third column illustrates how different the relationship of the two value estimates are in this failure case (from both successful and expert-specified GVFs).

Over the course of successful runs, GVFs found by MGD do not look exactly like the echo GVFs introduced in Section \ref{monsoon:intro}. Some characteristics are similar: i.e., one of the policies approaches $\pi \approx [1,0]$; however, the other policy $\pi \approx [0.8, 0.2]$ looks quite different from the deterministic policies. Even when the \emph{value estimates} output by the self-supervised GVFs are similar to those from expert-specified GVFs, the cumulant and policies can be quite different.

For the best parameter settings in our sweep, one of 30 independent trials failed, achieving an average reward per-step of 0.5 (note this performance is similar to that of the agent with only environmental observations). For this failed run, the learned policy and cumulant do not fit the categorisations of cumulants or policies learned in successful trials. Importantly, the learned value estimates presented to the control agent as features do not share the same cyclic values that capture the underlying seasons of the environment. From this failed run, we can see that simply adding \textit{any} prediction does not enable the agent to solve the problem: in successful trials, the policies and cumulants learned by MGD are meaningful and specific to the environment. These specific cumulants and policies are what enable the agents to solve the problem.

The policies and cumulants found through meta-gradient descent make sense when comparing to predictions chosen by domain experts. Importantly, Not just any prediction will do: we observed that successful runs can be categorised into particular policies and cumulants learned, and that the run that was unable to improve upon random behaviour learned a policy and cumulant that falls outside of the

\subsection{Learning to Specify GVFs in Frost Hollow}

\begin{figure}[h!]
    \centering
    \includegraphics[height=4cm, width=\linewidth, keepaspectratio]{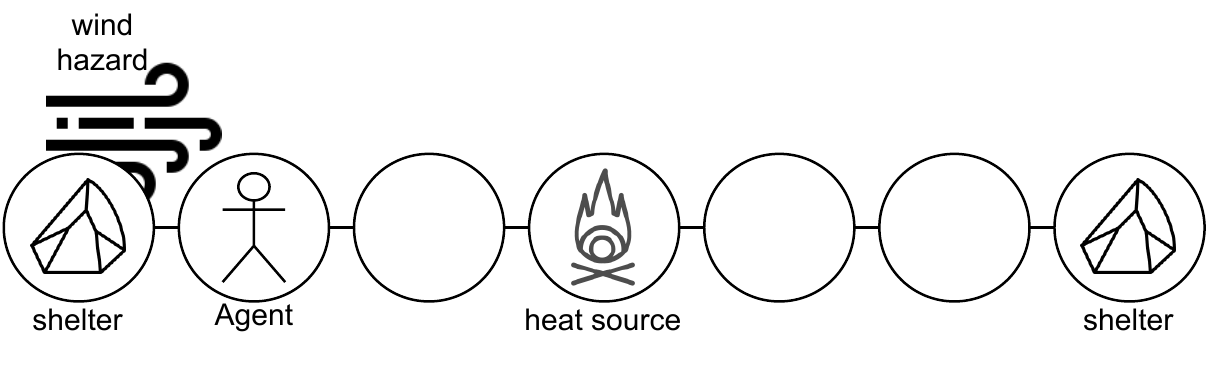}
    \caption{A depiction of the frost hollow problem. Frost hollow is a  linear walk where an agent collects a unit of heat by standing next to a fire in the center state. Once the agent accumulates 12 units of heat, it receives a reward of 1. Every 8 time-steps, a wind hazard gusts for two consecutive time-steps, removing all of the agent's accumulated heat if it is exposed to the hazard. To avoid losing its heat, the agent can take shelter in either of the end states. On each time-step, the agent observes its own location, the amount of heat it has accumulated, and whether the wind hazard is present.}
    \label{fig:frost-hollow}
\end{figure}

The previous example explored whether using MGD an agent could learn to specify predictions in order to resolve the partial-observability of its environment. In this section, we add two additional complications: 1) instead of a linear control agent, we use a more complex function approximator; and 2) we use a domain where the reward is sparse, thus complicating the GVF specification process. Frost Hollow (depicted in Figure \ref{fig:frost-hollow}) \citep{butcher2022pavlovian,brenneis2021assessing} was first proposed as a joint-action problem where two agents attempt to cooperatively solve a control problem: a prediction agent passes a cue to a control agent based on a learned prediction; using this cue as an additional input, the control agent takes an action. Frost Hollow was designed as an environment where predictive inputs are used to augment a control-agent's inputs,  making it well suited for assessing whether via MGD an agent can independently choose what GVFs to learn in order to improve performance on a control task.

\begin{figure}[h!]
    \begin{subfigure}[b]{.55\textwidth}
        \centering
        \includegraphics[height=1.5in, width=\linewidth, keepaspectratio]{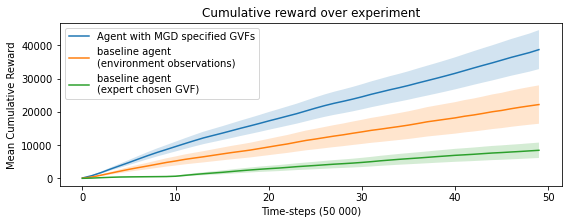}
        \caption{Mean cumulative reward over the entire duration of the trials in Frost Hollow. Assuming an agent follows the optimal policy deterministically for the total 2.5 milllion time-steps the maximum possible reward is $125000$ Error bars are the standard error.}
    \label{fig:fh:cumreward}
    \end{subfigure}
    \hfill
    \begin{subfigure}[b]{.4\textwidth}
        \begin{center}
            \begin{tabular}{c|c}
                \textbf{Agent Type} & \textbf{Cumulative Reward:} \\ 
                & \textbf{Evaluation Phase} \\
                \hline
                baseline & \\ 
                (environment obs) &  $7 \pm 2.9$  \\
                \hline
                baseline & \\
                (expert chosen GVF) & $3.3 \pm 1.6$\\
				\hline
                Agent with MGD & \\
                specified GVF & $18.7 \pm 4.2$ \\
            \end{tabular}
        \end{center}
        \caption{Cumulative reward and standard error of the mean during final 1000 evaluation steps for best configuration of each agent. Maximum possible cumulative reward is 50.}
    \label{fig:fh:table}
    \end{subfigure}
    \caption{Mean cumulative reward for each agent in the Frost Hollow, averaged over 30 trials.}
\label{fig:fh:results}
\end{figure}

\begin{figure}[h!]
\begin{subfigure}[b]{\linewidth}
 \includegraphics[width=\linewidth,keepaspectratio]{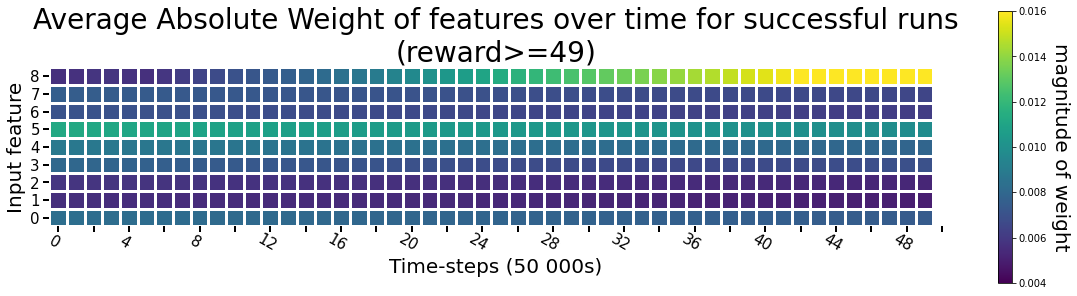}
\centering
\end{subfigure}
\begin{subfigure}[b]{\linewidth}
\includegraphics[width=\linewidth,keepaspectratio]{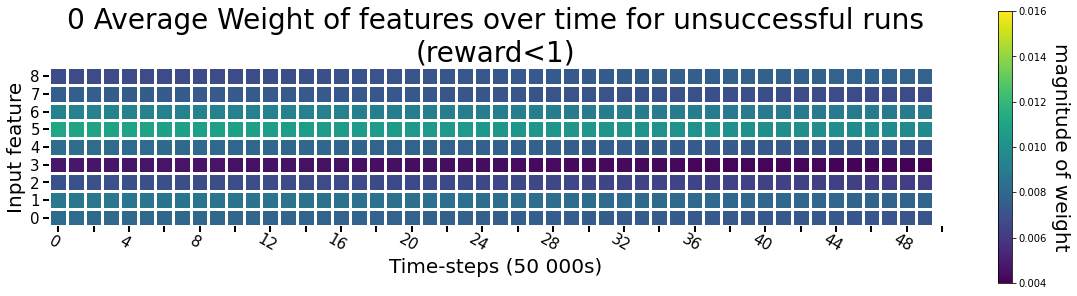}
  \centering
\end{subfigure}
  \caption{A depiction of the weights of the meta-weights $\omega_\text{c}$ for cumulants learned. Inputs $0 - 6$ are a binary encoding of the agent's location in the linear walk. Input 7 is a binary feature that encodes whether the wind hazard is present. Input 8 is the accumulated heat.}
  \label{fig:fh:cumulants}
\end{figure}

While simple, Frost Hollow poses a difficult learning problem. The reward is sparse: an agent can only observe a reward after successfully accumulating heat and dodging regular hazards. It takes at a minimum 50 time-steps, or 5 successive cycles of dodging the hazard successfully, before the agent can acquire a single reward. While the hazard itself is observable to the agent when active, the agent must pre-emptively take shelter before the hazard's onset in order to avoid losing its accumulated heat. All of this must be learnt from a sparse reward signal. Learning an additional GVF and using its estimate as an additional predictive feature can enable both humans, and value-based agents to successfully gain reward \citep{butcher2022pavlovian,brenneis2021assessing}.

In the frost-hollow setting, the control agent receives a single \textit{on-policy} prediction as an additional input feature: the GVF is conditioned on the agent's behaviour rather than a policy specified by MGD. The discount $\gamma$ is a constant valued $0.9$, following \cite{butcher2022pavlovian}. We compare the meta-gradient learning process to two baselines: 1) where the control agent receives a expert defined GVF, as specified in \cite{butcher2022pavlovian}; and where an agent that receives only the environment observations---no additional predictive features or information. In this setting we use a DQN agent for the control learner \citep{mnih2015human} adapted from Dopamine \citep{castro18dopamine}. We train all agents for 249 000 time-steps, and evaluate their performance on the final 1000 time-steps. During the evaluation phase the $\epsilon$ is set to 0.01, limiting non-greedy actions. All reported results are averaged over 30 independent trials. See Appendix B for additional experiment details.

In Figure \ref{fig:fh:table}, we report the average cumulative reward during the final evaluation steps. If an agent is deterministically following the optimal policy during the evaluation phase, the maximum possible cumulative reward is 50. In \cite{butcher2022pavlovian}, the best performing agent with a highly specialised representation was able to achieve a cumulative reward of around 40; however, many of the agents with hand-selected predictions introduced in \cite{butcher2022pavlovian} received a cumulative reward of approximately 30. 

In our experiments, the baseline agent without any additional predictions achieves an average cumulative reward of 7. By adding an additional predictive input feature that is specified by MGD, we are able to achieve an average cumulative reward of 18.7. Interestingly, the MGD agent learned to specify a cumulant that is different from those chosen in \cite{butcher2022pavlovian}. Successful runs learn a cumulant that predominantly weights the accumulated heat, input 8 (as depicted in Figure \ref{fig:fh:cumulants}). In Figure \ref{fig:fh:cumulants} we report the average meta-weights specifying the cumulant over the course of the entire experiment. In \cite{butcher2022pavlovian}, the expert chosen prediction is of the oncoming hazard: input 7. There is a logic to the prediction selected by MGD: the heat an agent accumulates is directly related to reward---reward is recieved after an agent acquires 12 heat. Moreover, the heat accumulated is indirectly related to the hazard: if the agent is unprotected before the hazard, it should anticipate a drop in it's accumulated heat. By predicting accumulated heat, the agent is dually capturing information about both the sparse reward signal and the original aspect of the environment that the expertly defined prediction sought to predict.

Interestingly, the baseline agent which used an expert-specified prediction performed less well than the MGD agent, receiving an average cumulative reward of 3.3. This is a revealing example: while the expert-specified GVF was well suited to the tabular setting explored in prior work \citep{butcher2022pavlovian}, its effectiveness as a predictive feature did not generalise to the function approximation setting. This highlights a challenge that we believe is present across domains and environmental settings. What predictive features might be useful to an agent is not only influenced by the environment, but also differences in state construction and the underlying learning method. Together, these factors all influence what GVFs may be useful for decision-making.

\section{Limitations \& Future Work }

In this paper we demonstrate for the first time that end-to-end learning of GVFs used as predictive inputs is possible. Moreover, we demonstrate this within the setting that GVFs were originally proposed: continual life-long learning. This has been an open challenge in GVF research since their introduction 10 years ago. However, this paper is not the final word on MGD discovery of GVFs. Assessing how well the method presented scales in environments with non-stationarity, or greater observational complexity is important future work. In this paper, we decided to fix the discount $\gamma$---the time-horizon over which a prediction is estimated. Previous work has enabled GVFs to be learned over many timescales at once, enabling inference over arbitrary horizons \citep{sherstan2020gamma}. Future work could explore how Sherstan {\em et al.}'s $\gamma$-net formulation  could improve the scalability and flexibilty of the meta-learning process we introduce. This paper makes progress in enabling agents to choose what to predict. How an agent may decide the number of predictions to learn remains an important open question. Similarly, how an agent might incrementally increase its capacity by adding new predictions during life-long learning remains to be explored. One possibility is to arrange predictions in multiple layers, similar to GVF networks: \citep{schlegel2021general}. Questions regarding GVF structure and scale are exciting open frontiers, and this manuscript provides a foundation that enables such questions to be asked in future work.

\section{Conclusion}
In this paper we introduced a process that enables agents to meta-learn the specifications of predictions in the form of GVFs. This process enabled agents to learn what aspects of the environment to predict while also learning the predictions themselves and learning how to use said predictions to inform action-selection. This meta-learning process was able to be implemented in an online, incremental fashion, making it possible for long-lived continual learning agents to self-supervise the specification and learning of their own GVFs. We found that an agent with no prior knowledge of the environment was able to select predictions that yielded performance equitable to, or better than agents using expertly chosen predictive features. Even in an environment with a sparse reward, an agent was still able to learn to specify useful predictions to use as additional features based on the control-learner's error. In sum total, this work provides an important step-change in the design of agents that use predictive knowledge.

It has long been suggested that predictions of future experience in the form of GVFs can provide useful features to support decision-making in computational reinforcement learning. Requiring system designers to re specify the GVFs for every permutation of an agent and its environment is a burden for applications of predictive features, and yet it is still the norm in both research and applied settings. Fundamentally, the requirement of designers to hand-specify GVFs prevents the development of fully independent and autonomous agents. Moreover, this has inhibited the application of GVFs, especially in long-lived continual learning domains that may exhibit non-stationarity. In this paper, we contributed a meta-gradient descent processes by which agents were able to find GVFs that improved decision-making relative to environment observations alone In doing so, we provide a research path for resolving an open challenge in the GVF literature that has existed for over a decade. 

\subsubsection*{Acknowledgments}

This work was supported in part by The Natural Sciences and Engineering Research Council of Canada (NSERC), the Canada CIFAR AI Chairs Program, the Alberta Machine Intelligence Institute (Amii), and the Canada Research Chairs program. AKK was supported by scholarships and awards from NSERC and Alberta Innovates.

\bibliography{main}
\bibliographystyle{collas2022_conference}
\vfill
\pagebreak
\appendix
\section{Experiment Details: Monsoon World}
\label{appendix:monsoon}

Monsoon world experiments ran for a total of one million time-steps. Each agent had a training phase of 990,000 time-steps. The agent's performance is evaluated during the final 1000 time-steps where $\epsilon$ is set to 0 and actions are chosen greedily so that we can compare average reward achieved from following the learned policies.

Monsoon world experiments are not run-time sensitive.

\subsection{Function Approximators and State Aggregation}

We use different function approximators to transform the given inputs to an \textit{agent-state} $\vs_t = \phi(\vo_t, \vv_t)$. Echo GVF estimates are in log-space (c.f. \cite{schlegel2021general} for more information on echo GVFs). 
Before using an echo GVF's estimate as inputs, we apply a log transformation to them (see Algorithm~\ref{alg:FAlog}). 

A simple state-aggregation algorithm, \FAagg is given in Algorithm~\ref{alg:FAagg}). We will abbreviate linear function approximation as \FAlin.

        \begin{algorithm}
                \caption{Log transform of prediction estimates}
            \begin{algorithmic}
            \State hello
                \State \# Where $\vv_{}$ holds the value estimate from $n$ GVFs.
                \State \# Where $\max\gamma_t$ is determined by the $g$ function.
                \Statex
                \Procedure{transform}{$\vv_{}$}
                    \State $\vv \gets \mathtt{clip}(\log(\vv_{})/\log(\max\gamma_t), 0,1)$
                    \State return $\vv_{}$
                \EndProcedure
            \end{algorithmic}
        \label{algx:FAlog}
        \end{algorithm}
        \begin{algorithm*}
            \caption{State aggregation of predictions, \FAagg}
            \begin{algorithmic}
                \State \# Where $\vv_{}$  are the value estimates from $n$ GVFs.
                \State \# Where $\evv^i < 10$
                \State \# Where \texttt{memsize} is the allocated length for the binary feature vector.
                \Statex
                \Procedure{state}{$\vv_{}$}
                    \State $\vs = \mathtt{zeros}(\texttt{memsize})$
                    \State $i \gets \lfloor\vv_{0} + \vv_{1}*10\rfloor$
                    \State $\evs^i = 1$
                    \State return $\vs$
                \EndProcedure
            \end{algorithmic}
        \label{alg:FAagg}
        \end{algorithm*}
  
We compared three agents in the Monsoon environment: the agent that learns a value solely over the observations, \obs; the agent that learns over both observations and expert-specified GVFs, \expert; and the agent that learns over both observations and GVFs discovered through meta-gradient descent, \mgd.

Each agent has a specific function-approximator for each of its component. 
All RL agents have a function approximator used by the control unit, $\phi$. A GVF-based agent also has a choice of function approximator for the GVF predictions, $\phi^\nu$. Finally, the meta-gradient agents may also have a function approximator used for the meta-parameter update.

Parameters were chosen by performing a sweep across different values, choosing the best performing combination for each agent during the final 1000 evaluation steps in the experiment.

\begin{figure}
    \begin{subfigure}[t]{.5\textwidth}
        \begin{center}
            \begin{tabular}{ l| c c c }
                Agent & $\phi$ & $\phi^\nu$ & $\phi^\omega$ \\
                \hline
                \obs & \FAagg & \FAagg & -\\
                \expert & \FAagg & \FAagg & -\\
                \mgd & \FAlin & \FAagg & \FAlin \\
            \end{tabular}
        \end{center}
    \caption{Function approximators used for each agent component.}
    \end{subfigure}
    \begin{subfigure}[t]{.5\textwidth}
        \begin{center}
            \begin{tabular}{ r| c c c c c }
                Agent & $\epsilon$ & $\alpha^Q$ & $\alpha^V$ & $\alpha^{\rho}$ & $\alpha^{c}$ \\
                \hline 
                 \obs & 0.1 & 0.01 & 0.1 & - & - \\ \hline
                 \expert & 0.1 & 0.01 & 0.1 & n/a & n/a \\ \hline
                 \mgd &  0.5 & 0.0001 & 0.1 & 0.001 & 0.1 \\ \hline
            \end{tabular}
        \end{center}
        \label{tab:monsoon-parameters}
        \caption{Hyperparameters for each agent.}
    \end{subfigure}
    \caption{Parameter settings for different agent configurations}
\end{figure}

\subsection{Meta-parameter Specification}

The GVF's policy $\pi$ is a fixed policy: the meta-weights determine the policy a GVF is conditioned on, but they are not a function of the observations: $\pi \gets \softmax(\vomega^\pi)$

The cumulant $c$ is a function of the observations, $c_t = \sigmoid(\vo_t^\top\vomega^c)$,
where $\vomega^c$ are the meta-weights for the cumulant, and $\vo_t$ is the present environment observation.

\section{Experiment Details: Frost Hollow}
\label{appendix:fh}

All independent runs lasted 2,500,000 time-steps. Each agent had a training phase of 2,499,000 time-steps, and an evaluation phase in the final 1000 time-steps where $\epsilon=0.001$ for $\epsilon$-greedy action selection.

\subsection{Function Approximators}

As in the Monsoon World experiments, the meta-weights that specify the cumulants are a linear combination of features and weights with no activation function. For a single GVF the cumulant at time $t$ is: $c_t = \vomega_t^\top\vo_t$.

The GVF in the frost-hollow experiment uses a \textit{bit-cascade} representation, as introduced in \citep{butcher2022pavlovian}.

The control agent uses an artificial neural network as its function approximator. In particular, \texttt{JaxDQNAgent} from \cite{castro18dopamine} and use the default parameter configurations for all control agents.

\subsection{Parameter Settings}

The DQN configuration uses the default parameter settings of \cite{castro18dopamine} for the \texttt{JaxDQNAgent}. 

Additional reported parameters were chosen by performing a sweep across different values, choosing the best performing combination for each agent during the final 1000 evaluation steps in the experiment.

\begin{table}[H]
    \centering
    \begin{tabular}{| l | c | c |}
         \hline
         Agent &   $\alpha^V$ &  $\alpha^Q$ \\
         \hline 
         \obs  & n/a & n/a \\ \hline
         \expert  & 0.001 & n/a  \\ \hline
         \mgd & 0.001 & 0.0001  \\ \hline
    \end{tabular}
    \caption{Best parameter settings for different agent configurations}
    \label{tab:fh:pararmeters}
\end{table}
\clearpage
\section{Detailed pseudo-code for Meta-gradient Reinforcement Learning}
\label{appendix:alg}
    \begin{algorithm*}
    \caption{A meta-gradient approach to self-supervised predictive Reinforcement Learning.} 

    \begin{algorithmic}
    \Statex
    \State \textbf{Choose hyperparameters:}
    \State Choose control agent's state function approximator $\phi$, learning rate $\alpha$, L2 regularizer $\lambda$ and exploration rate $\epsilon$.
    \State Choose GVF state function approximator $\phi^\nu$, eligibility trace factor $\lambda^\nu$ and learning rate $\alpha^\nu$ for GVF updates.
    \State  Choose $\alpha^c$, $\alpha^\rho$ for meta-learning updates.
    \Statex
    \State \textbf{Initialise:}
    \State Initialise Q-learning weights $\vw_0$
    \ForAll{$i \in$ GVFs}
        \State Initialise GVF prediction $\evv^i_0$
        \State Initialise weights $\vnu^i_0$ and $\vomega^i_0$ 
    \EndFor
    \Statex
    \State \textbf{BEGIN}
    \State Receive initial observation $\vo_0$
    \State Calculate initial agent-state vector $\vv_0$
    \Statex
    \For{$ t=1 \dots $}
    \State \# \textbf{Control Agent}
    \State Compute agent-state $\vs_{t-1}\gets\phi(\vo_{t-1},\vv_{t-1})$
    \State Use action-values $Q(\vs_{t-1}, a; \vw_{t-1})$ and control policy $\pi$ to choose action $a_{t-1}$
    \State Take action $a_{t-1}$, observe $r_{t}$ and $\vo_t$
    \Statex
    \For {$i\in$ GVFs}
        \State \# \textbf{Meta-gradient Update}
        \For {each parameterized GVF component $j\in \{c, \rho\} $}
            \State Update meta-weights $\vomega^j_{t} \gets\vomega_{t-1} - \alpha^j\nabla_{\!\vomega^{j}}\Ls_t$
        \EndFor
        
        \Statex
        \State \# \textbf{GVF Value Update}
        \State Calculate $\rho^i_t$, $c^i_t$, $\gamma^i_t$ using $\vomega^i_t$
        \State Update GVF weights $\vnu^i_t$ according to GVF learning procedure.
    \EndFor
    \Statex
    \State \# \textbf{Control Agent Update}
    \State Compute current agent-state $\vs_t \gets \phi(\vo_t, \vv_t)$
    \State Calculate Q-learning TD error $\delta_t \gets r_t + \gamma\max_a Q(\vs_{t}, a) - Q(\vs_{t-1},a_{t-1})$
    \State Update agent weights $\vw_t$ according to control agent learning procedure.
    \EndFor
    \end{algorithmic}
    \label{alg}
    \end{algorithm*}
 
\end{document}